\documentclass[runningheads]{llncs}
\usepackage[T1]{fontenc}
\usepackage{cite}
\usepackage{amsmath,amssymb,amsfonts}
\usepackage{graphicx}

\usepackage{algorithm}
\usepackage{algpseudocode}

\usepackage{multirow}
\usepackage{array}
\usepackage{tabularx}
\usepackage{makecell}
\usepackage{booktabs}
\usepackage{cellspace}
\usepackage{hyperref} 

\hypersetup{
    colorlinks=true,    
    linkcolor=blue,     
    citecolor=blue,      
    urlcolor=blue       
}
\usepackage{verbatim}
\usepackage{algorithm}
\usepackage{algorithmicx}
\usepackage{algpseudocode}
\usepackage{textcomp}
\usepackage{xcolor}
\usepackage{orcidlink}
\begin{document}
\title{
  \raisebox{-0.15em}{\includegraphics[height=1em]{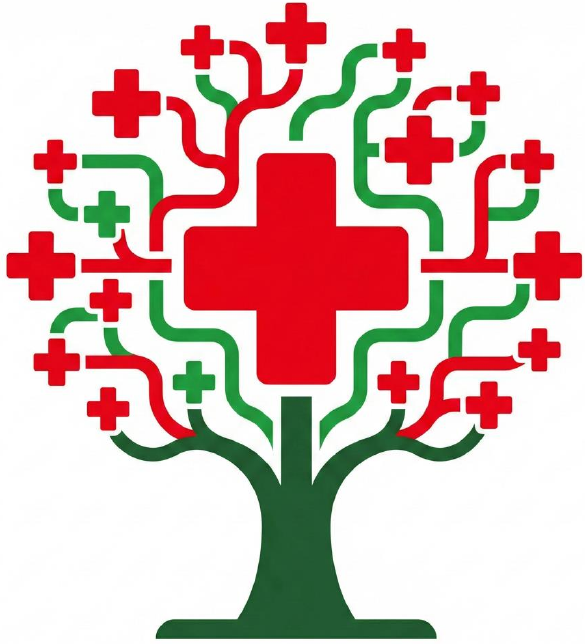}}%
  \hspace{0.5em}
  QM-ToT: A Medical Tree of Thoughts Reasoning Framework for Quantized Model
  }

\author{
Zongxian Yang\inst{1}\orcidlink{0009-0002-6230-4301} \and
Jiayu Qian\inst{1}\orcidlink{0009-0007-7148-2843} \and
Kay Chen Tan\inst{2}\orcidlink{0000-0002-6802-2463} \and
Hau-San Wong\inst{3}\orcidlink{0000-0002-1530-7529} \and
Yulong Chen\inst{1}\orcidlink{0000-0003-1821-3330} \and
Haoyu Zhang\inst{3}\orcidlink{0009-0004-0616-0096} \and
Zhi-An Huang\inst{1}\orcidlink{0000-0001-9974-148X}\thanks{Corresponding author.}
}

\institute{
City University of Hong Kong(Dongguan), Dongguan, China \inst{1} \and
Hong Kong Polytechnic University, Hong Kong, China \inst{2} \and
City University of Hong Kong, Hong Kong, China \inst{3} \\[0.5em]
\email{huang.za@cityu-dg.edu.cn}
}
%
%
\maketitle              

\begin{abstract}
Large language models (LLMs) have achieved substantial progress in biomedical question answering. However, real-world medical applications often operate under resource-constrained settings, where model quantization is a practical necessity for local and privacy-preserving deployment. In such settings, the inherent complexity of clinical reasoning further amplifies the performance degradation of quantized LLMs. To address these issues, we propose Quantized Medical Tree of Thought (QM-ToT), a structured tree-of-thought reasoning framework that supports feedback-guided clinical reasoning. QM-ToT decomposes complex medical problems into hierarchical reasoning paths and selects answers under dual-evaluation feedback.  This framework facilitates substantial performance improvements in INT4-quantized models on the challenging MedQA-USMLE dataset. Specifically, we demonstrate a remarkable accuracy increase from 34\% to 50.25\% for the LLaMA2-70b model and from 58.77\% to 69.49\% for LLaMA-3.1-8b. Besides, we also proposed an effect data distillation method based on QM-ToT. Besides, we propose Reflection-ToT, a data distillation method based on QM-ToT. Using 1,000 questions, Reflection-ToT achieves 66.44\% accuracy on LLaMA3.1-8B, compared with 65.01\% from QwQ-based CoT distillation using 10,178 questions and 64.73\% from a QwQ-based matched-pair setting with 1,514 pairs.
The distilled data has been published\footnote{https://anonymous.4open.science/r/QM-ToT-DistillationData-5E0E}. 
This work demonstrates the potential of structured tree-based reasoning with evaluator feedback for improving biomedical question answering under low-precision deployment, while also highlighting the need to account for evaluator dependence and inference cost.
\end{abstract}
%
%
%


\section{Introduction}
Large language models (LLMs), trained on massive text corpora, have revolutionized numerous Artificial Intelligence (AI) tasks, from recognition and classification to prediction~\cite{teacher1,teacher2,teacher3}. Their ability to comprehend human instructions and solve complex problems has opened exciting new avenues in various domains. Although existing LLMs have shown promise in general medical language understanding, their performance on specialized biomedical benchmarks remains to be improved~\cite{ref1}.

This performance gap is primarily attributable to two key challenges. First, biomedical text analysis and interpretation demand a high level of reasoning complexity. LLMs must not only decipher intricate medical terminology but also generate accurate and clinically relevant insights conforming to professional standards and guidelines. Even state-of-the-art models like OpenAI-o1~\footnote{https://openai.com/o1/} require specialized strategies to enhance their reasoning capabilities in this domain~\cite{ref2}, highlighting the need for innovative solutions.

Second, the sensitive nature of medical data raises significant privacy and security concerns~\cite{ref3}. Data breaches in healthcare can have severe consequences, potentially compromising patient privacy and safety. Deploying large models locally~\cite{ref4} offers a promising solution to mitigate these risks. Quantization techniques, which minimize memory overhead, enable deployment on resource-constrained devices with reduced operational costs~\cite{ref5}. However, quantization could compromise model performance and output quality~\cite{ref6}, raising a significant challenge for practical application. Establishing such local quantized systems ensures the privacy of patient data while maintaining high-quality clinical insights.

\begin{figure}
    \centering
    \includegraphics[width=0.8\linewidth]{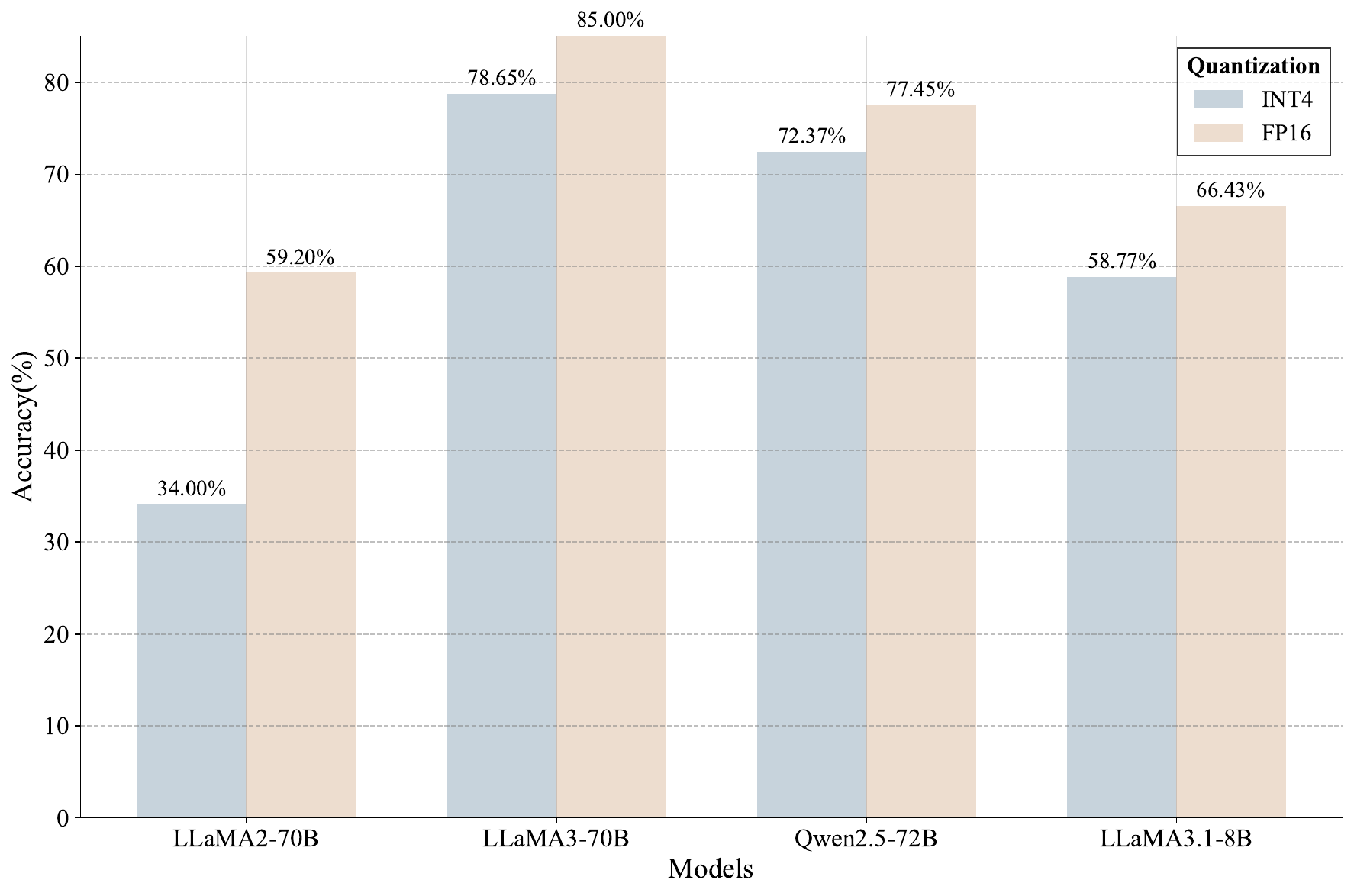}
    \caption{FP16 vs INT4 quantization performance comparison. Performance gap between FP16 and INT4 quantization for LLaMA3-70b, LLaMA2-70b, Qwen2.5-72b, and LLaMA3.1-8b models on the MedQA-USMLE dataset.}
    \label{fig:1}
\end{figure}

\begin{figure}[t]
    \centering
    \includegraphics[width=0.5\linewidth]{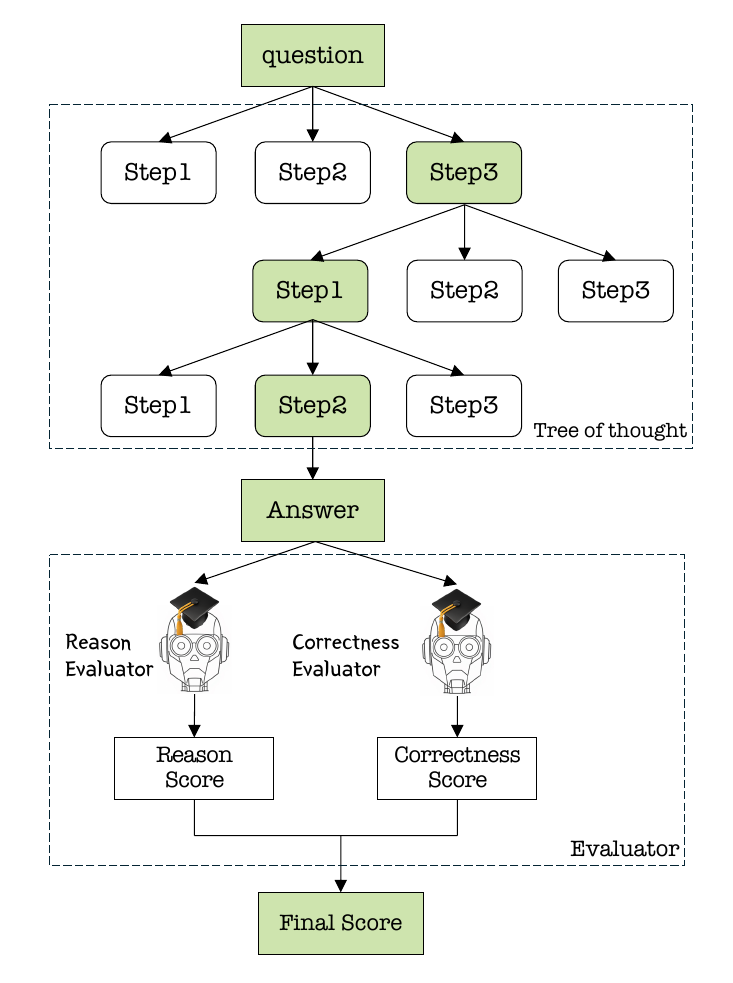}
    \caption{Tree-based Reasoning and Dual-Evaluation Workflow. This diagram showcases how the final score is calculated. Starting with a question, the workflow expands into a tree structure where each path can branch into multiple sub-paths, eventually leading to an answer. The answer and the chain of thought is then evaluated by two specialized evaluators - one for reasoning and one for medical fact correctness. These evaluations combine to produce a final quality score.}
    \label{fig:2}
\end{figure}

\begin{figure}[t]
    \centering
    \includegraphics[width=0.5\linewidth]{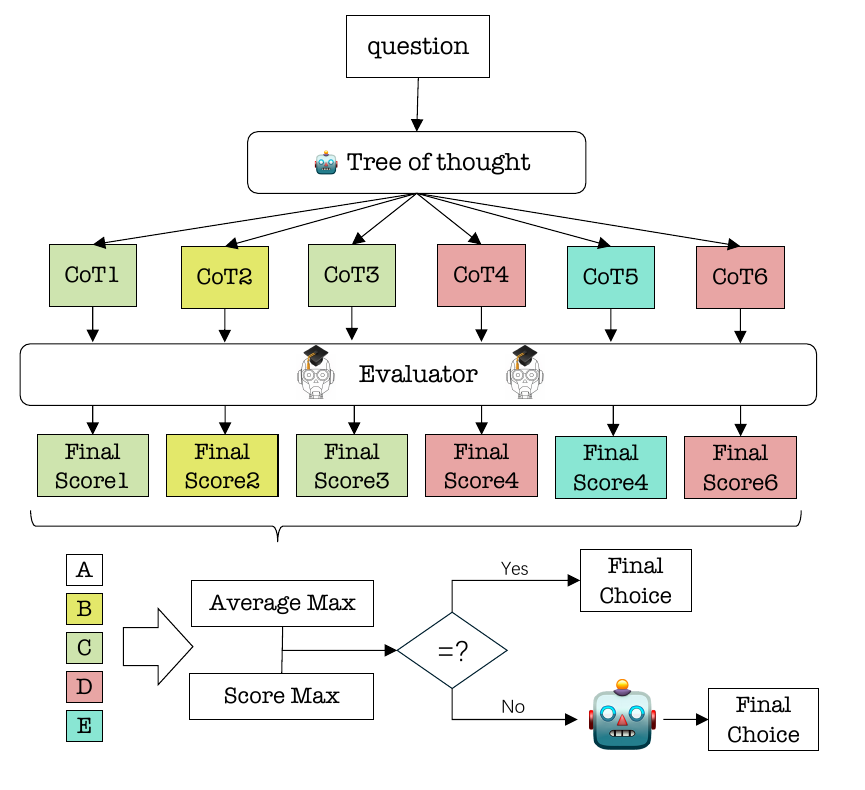}
    \caption{QM-ToT decision workflow. This workflow diagram illustrates the decision-making process of the QM-ToT framework, from initial question to final answer generation. The process begins with a question that feeds into a tree of thought system, generating different chain of thought. These chains are then evaluated by an Evaluator component, which assigns Final Scores to each path. The framework then compares two metrics: the option with the highest average score (Average Max) and the option with the highest individual Final Score (Score Max). If these two selections match, it directly determines the Final Choice. If they differ, the model initiates a re-evaluation process to reach the Final Choice. }
    \label{fig:3}
\end{figure}

The performance degradation caused by quantization is particularly acute in medical applications where accuracy is paramount. Our empirical analysis as shown in Fig.~\ref{fig:1} reveals a significant performance drop when quantizing models from FP16 to INT4 on the United States Medical License Exams (MedQA-USMLE) dataset~\cite{ref34}. This underscores the urgent need for strategies to maintain model robustness while enabling efficient quantization.
To address these challenges, we propose QM-ToT, a structured tree-of-thought reasoning framework for quantized LLMs in resource-constrained clinical question answering. As shown in Fig.~\ref{fig:2} and Fig.~\ref{fig:3}, our framework decomposes medical problem-solving into discrete paths, forming a ToT structure. The evaluator module provides internal feedback on factual accuracy and logical validity, enabling the framework to ground its solution search. The experimental results of the MedQA-USMLE dataset demonstrate that our approach significantly boosts accuracy compared to chain of thought with self-consistency (CoT-SC): Qwen2.5-72b from 70.25\% to 74.25\%, LLaMA-2-70b from 27.75\% to 50.25\%, and LLaMA-3.1-8b from 59.19\% to 69.49\%. Additionally, we categorize the dataset into easy, medium and hard levels to validate the effectiveness of both the ToT framework and the solution evaluator. This analysis provides interpretable evidence of our approach's efficacy across varying difficulty levels, illuminating the underlying mechanisms of the ToT method.

\begin{figure}[t]
    \centering
    \includegraphics[width=0.5\linewidth]{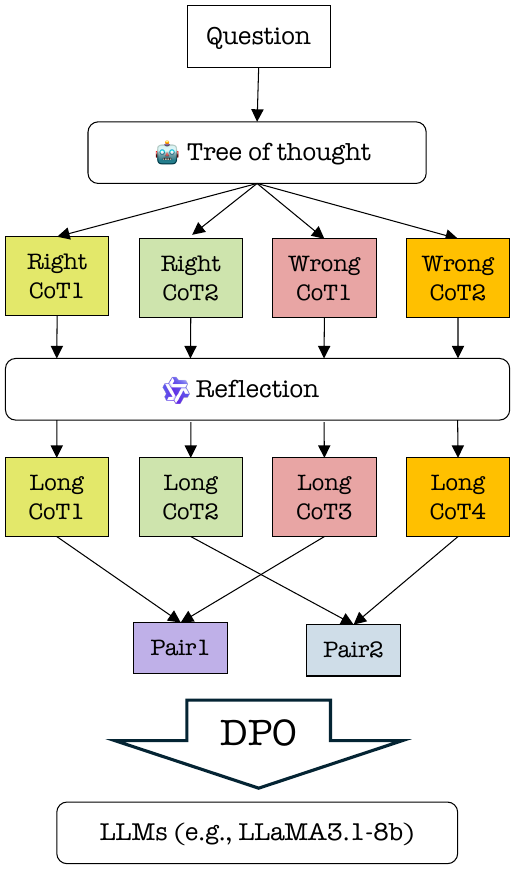}
    \caption{Reflection-ToT: a data distillation method driven by ToT. Short CoT generated by ToT without reflection is refined by the teacher model (eg. Qwen2.5-72b or LLaMA3-70B) to produce long CoT. Correct and incorrect long CoT are randomly paired to form Direct Preference Optimization (DPO) training data.}
\label{fig:Reflection-ToT}
\end{figure}

We also investigate the potential of ToT algorithms for constructing synthetic medical reasoning data through Reflection-ToT, as shown in Fig.~\ref{fig:Reflection-ToT}. This pipeline refines short CoT generated by ToT into reflection-augmented long CoT for student-model training. In our preliminary experiment, Reflection-ToT achieves 66.44\% accuracy using 1,000 questions and 1,514 preference pairs. This result is higher than QwQ-based CoT distillation using 10,178 questions (65.01\%) and a QwQ-based matched-pair setting with 1,514 pairs (64.73\%).

We summarize our contributions as follows:

\begin{itemize}
\item QM-ToT consists of two core components: a path-based ToT planner and a dual-evaluator module. The planner decomposes complex medical queries into hierarchical reasoning paths, while the evaluator provides feedback-driven quality assessment. This structured reasoning framework achieves an average accuracy gain of 9.14\% over CoT-SC on INT4-quantized models.

\item The novel ToT-driven data distillation pipeline named Reflection-ToT converts short CoT to long o1-style CoT with reflection. Using only 9.8\% of the question, we achieved obvious improvement compared to the traditional CoT distillation strategy from o1-like model QwQ~\footnote{https://qwenlm.github.io/zh/blog/qwq-32b-preview/} (a recently released model with output alignment to OpenAI-o1). This demonstrates the potential of the ToT method in medical data distillation tasks. 

\item  We study the adaptation of tree-based reasoning to medical question answering under quantized deployment. While ToT has shown promise in domains such as mathematical reasoning~\cite{ref7} and coding~\cite{ref8}, our experiments examine its behavior in biomedical multiple-choice reasoning with dual evaluator feedback.
\end{itemize}

\noindent\textbf{Data/Code Availability:}  The public repository link will be provided in the camera-ready version upon acceptance.

\section{Related Works}

\subsection{Artificial Intelligence Biomedical Tasks}

Biomedical tasks can be broadly categorized into two main types~\cite{ref13}: discriminative and generative tasks. Typical discriminative tasks include question answering, feature extraction~\cite{teacher4}, disease diagnosis~\cite{teacher5}, classification~\cite{teacher6}, multi-objective optimization\cite{chen5}, concepts predictions\cite{chen6}, among others\cite{11125826}. In contrast to discriminative tasks, generative tasks require models to produce fluent and contextually appropriate text based on specific inputs. The tasks include medical text summarization~\cite{ref14,ref15}, medical text generation~\cite{ref16,yang2025med}, and text simplification~\cite{ref17}. LLMs have been widely adopted in medical research and applications due to their robust reasoning capabilities. A common approach is to integrate general-purpose AI to assist healthcare professionals in their work~\cite{jiang2025tc}. Another approach involves developing specialized medical LLMs through pre-training or fine-tuning on extra medical datasets, as exemplified by MedPaLM-2~\cite{ref11} and Med-Gemini~\cite{ref12}. More broadly, advanced computational models have also been explored in adjacent application domains such as traffic prediction, edge networks, and biomedical classification, reflecting the increasing role of data-driven modeling in complex real-world systems~\cite{ali2026advanced,yin2018edge}.

\subsection{Tree of Thought Reasoning}

Tree of thought is a cognitive framework that leverages LLMs to emulate human reasoning and problem-solving processes. In the 1950s, Newell~\cite{ref28} conceptualized problem-solving as a search process through combinatorial problem spaces, represented in a tree structure. In 2023, Yao~\cite{ref7} pioneered the application of tree search as a problem-solving paradigm for LLMs. Subsequently, the ToT approach demonstrated its efficacy, exhibiting remarkable performance in various domains, including mathematics~\cite{ref7} and coding~\cite{ref8}. Additionally, ToT has been recognized as an effective data augmentation technique for generating domain-specific training data, thereby enhancing model performance~\cite{ref29}.

\subsection{The Rise of LLM-as-a-Judge}

Recent LLMs, such as GPT-4~\cite{ref20} and Qwen2.5~\cite{ref21}, have demonstrated exceptional capabilities in instruction compliance, query comprehension, response generation and causal Learning~\cite{chen4}. These advancements have led to the emergence of the ``LLM-as-a-judge" paradigm~\cite{ref22}. This framework leverages powerful LLMs to evaluate, rank, and select from a pool of candidates. This approach significantly addresses the limitations of conventional evaluation methodologies and has been extensively implemented across the LLMs lifecycle~\cite{ref23}, encompassing tasks such as alignment~\cite{ref24}, retrieval~\cite{ref25}, and reasoning~\cite{ref26}. This paradigm further endows LLMs with advanced capabilities like self-evolution~\cite{ref27}.

\subsection{Quantization for Efficient LLM Deployment}

While LLMs demonstrate remarkable performance across numerous tasks, their reasoning processes are computationally demanding. To address this challenge, researchers have developed quantization techniques that convert floating-point model parameters into integer representations, thereby reducing storage requirements and accelerating computational processes. Prominent quantization approaches include GPTQ~\cite{ref18}, AWQ~\cite{ref19}, and GGUF\footnote{https://github.com/ggerganov/ggml}. While quantized models may exhibit some performance degradation, quantized high-parameter models typically outperform their unquantized smaller counterparts~\cite{ref6}, despite requiring comparable GPU memory footprints. Consequently, the deployment of quantized high-parameter models in practical applications has emerged as a compelling solution. Related studies on quantized sampled-data systems, secure consensus control, and lightweight authentication further highlight the importance of robustness and efficiency in resource-constrained intelligent
  systems~\cite{samy2025secure,sharafian2025consensus,haider2024advanced}.

\section{Method}
\subsection{ToT for Quantized Medical Reasoning}

QM-ToT follows a structured reasoning loop: (i) \textbf{Planning} through path decomposition, (ii) \textbf{Execution} via thought generation, and (iii) \textbf{Reflection} through dual-evaluation feedback. This enables quantized LLMs to explore diverse reasoning pathways through a controlled tree-search procedure, mitigating performance degradation through structured, feedback-guided search.
Formally, the generation of a reasoning path can be defined as,
\begin{equation}
\label{eq:eq1}
s_{0}=D_{\pi}(Q),
\end{equation}
where $Q$ represents the medical problem description with specific problem statements and requirements, $D_{\pi}$ denotes the path decomposition function parameterized by the LLM $\pi$, and $s_0$ is the initial reasoning path.

Built upon the initial path, the subsequent reasoning paths can be achieved as,
\begin{equation}
\label{eq:eq2}
s_{1}=D_{\pi}(Q+s_0),
\end{equation}

where $s_{1}$ represents a reasoning path derived from the problem description $Q$ and the preceding path $s_0$. Note the inclusion of both $Q$ and $s_0$ as input to $D_\pi$, ensuring the LLM retains context of the original problem.

This iterative process continues, ultimately leading to a final solution:
\begin{equation}
\label{eq:eq3}
a_Q=s_{k}=D_{\pi}(Q+\sum_{t=0}^{k}s_t),
\end{equation}

where $a_Q$ represents the final solution to problem $Q$, $s_{k}$ represents the cumulative sequence of reasoning paths, constrained by a maximum length $k$. This formulation emphasizes the dependence of each path on the entire reasoning chain.

QM-ToT employs an evaluation model $W$ (where DeepSeek-V3~\footnote{https://github.com/deepseek-ai/DeepSeek-V3} is used here), also driven by the LLM $\pi$, to execute these paths and assess their validity in solving the problem. Algorithm 1 provides pseudo-code illustrating the ToT framework.

\begin{algorithm}[t]
\caption{QM-ToT Reasoning with Maximum Length}
\begin{algorithmic}[1]
\Statex \textbf{Input} $Q$: Medical problem description, $k$: Maximum length of reasoning paths
\Statex \textbf{Output} Solution set $a_Q$
\State $S \gets \{\}$ \hspace{1em} \Comment{Initialize the set of reasoning paths}
\State $s_0 \gets D_{\pi}(Q)$ \hspace{1em} \Comment{Generate the initial path}
\State $S.\text{add}(s_0)$
\State $i \gets 1$ \hspace{1em} \Comment{Path index}

\While{\textbf{not} $W(Q, S)$ \textbf{and} $i \leq k$} \hspace{1em} \\\Comment{While the problem is not solved and within max length}
\hspace{1em} \For{\textbf{each} possible next path $s_{i}$ from $D_{\pi}(Q, S)$}
\hspace{2em} \State $S' \gets S \cup \{s_{i}\}$ \hspace{1em} \Comment{Create a new branch}
\hspace{2em} \If{$W(Q, S')$}
\hspace{3em} \State $S \gets S'$ \hspace{1em} \Comment{Update current reasoning paths}
\hspace{3em} \State $i \gets i + 1$ \hspace{1em} \Comment{Increment the path index}
\hspace{3em} \State \textbf{break} \hspace{1em} \Comment{Move to the next level of the tree}
\hspace{2em} \Else
\hspace{3em} \State $S' \gets S' \setminus \{s_{i}\}$ \hspace{1em} \Comment{Backtrack if invalid paths}
\hspace{2em} \EndIf
\hspace{1em} \EndFor
\EndWhile \\

\If{$W(Q, S)$}
\hspace{1em} \State \Return the final path in $S$ as $a_Q$
\Else
\hspace{1em} \State \Return \textbf{Failure to Solve} \hspace{1em} \Comment{If $k$ is exceeded}
\EndIf

\end{algorithmic}
\end{algorithm}

\subsection{Result Evaluator}

In medical question-answering scenarios with multiple-choice questions, selecting the best answer from several generated options requires careful evaluation. While self-consistency through statistical aggregation is a common approach, it can be unreliable, especially when misleading information is present in the query. To mitigate this, we introduce a scoring mechanism that leverages both logical coherence and medical accuracy. We employ a specialized scoring model (i.e., DeepSeek-V3) to assess each generated reasoning chain, producing two scores: a reasoning coherence score ($r$) and a medical correctness score ($c$). These are combined into a final score ($fs$) for each chain using a weighted exponential average:
\begin{equation}
fs = \alpha \cdot \exp(r) + (1-\alpha) \cdot \exp(c)  \quad\quad 0 \leq \alpha \leq 1.
\label{eq:alpha}
\end{equation}

For each multiple-choice option $x \in {A, B, C, D, E}$, we compute both the average final score ($\operatorname{Avg}(fs)_x$) across all reasoning chains leading to that option and the maximum final score ($\max(fs)_x$) among those chains.

If the option with the highest average score and the option with the highest maximum score are the same, that option is selected as the final answer ($a$):

\begin{equation}
a = \underset{x}{\operatorname{argmax}}( \operatorname{Avg}(fs)_x). 
\end{equation}

If these options differ, a judge model ($J_{\pi}$), parameterized by the LLM $\pi$, compares the two options and selects the final answer:

\begin{equation}
a = J_{\pi}(\underset{x}{\operatorname{argmax}}( \operatorname{Avg}(fs)_x), \underset{x'}{\operatorname{argmax}}( \max(fs)_{x'})).
\end{equation}

This two-stage evaluation process, combining a specialized scoring model with a judge model, improves the robustness and accuracy of answer selection in complex medical question-answering tasks.

\subsection{Reasoning and Bias Boundaries}
\label{sec:rb}
To analyze the effectiveness of ToT reasoning on the MedQA dataset, we adapt the concept of reasoning boundary (RB)~\cite{ref33}. RB quantifies the maximum difficulty level at which a language model can effectively reason. Mathematically, for a model $m$ and task $t$, RB is defined as:

\begin{equation}
    \mathcal{B}_{K_1}(t|m) = \sup_{d} \{ d \mid \text{Acc}(t|d, m) \geq K_1 \},
\end{equation}

where $\text{Acc}(t|d, m)$ is the model's accuracy on task $t$ with difficulty level $d$, and $K_1$ is a predefined accuracy threshold. In the context of multiple-choice medical questions (normally with 4-5 options), we introduce the concept of a bias boundary ($\Omega$), representing the accuracy achievable through random guessing (20-25\% for MedQA). When a model's CoT-SC performance drops below the random baseline, it is viewed as a sign of model bias  rather than an indicator of its reasoning capability.

\begin{figure}[t]
    \centering
    \includegraphics[width=0.7\linewidth]{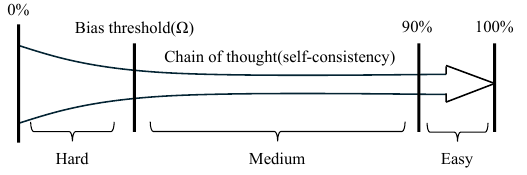}
    \caption{Difficulty classification of the dataset based on CoT-SC accuracy}
    \label{fig:4}
\end{figure}

As shown in Fig.~\ref{fig:4}, we categorize the questions into three difficulty levels based on the CoT-SC performance:

\noindent\textbf{Easy Level} ($\mathcal{B}_{\text{Acc}\geq90\%}(t|m)$): Questions where $\text{Acc}(t|m) \geq 90\%$. The model demonstrates consistent high performance, indicating reliable reasoning.

\noindent\textbf{Medium Level} ($\mathcal{B}_{\Omega<\text{Acc}<90\%}(t|m)$): Questions where $\Omega < \text{Acc}(t|m) < 90\%$. The model performs above chance but below high reliability, suggesting meaningful but imperfect reasoning. The model may need multiple attempts or additional prompting to produce correct answers.
                        
\noindent\textbf{Hard Level} ($\mathcal{B}_{\text{Acc}\leq\Omega}(t|m)$): Questions where $\text{Acc}(t|m) \leq \Omega$. Performance at or below the random baseline indicates reasoning fails or bias dominates. This hierarchical categorization enables systematic analysis of model capabilities across different question complexities.

\subsection{Data Distillation for Data Generation}
One feature of ToT is that the model explores solution paths in different directions, thus generating both correct and incorrect CoT simultaneously. This perfectly meets the requirements of DPO~\cite{DPO} training. We set up a reflector ($R$) driven by model $µ$. Thus, the distilled dataset ($T$) can be defined as:
\begin{equation}
T=\sum_{Q=1}^{n}R(M(a_Q)),
\end{equation}
where $Q$ represents the question in MedQA-USMLE and $M$ represents the random match policy to create the right-against-wrong sample pairs.

\section{Experiments}

We evaluated our approach on the test set of MedQA-USMLE with 1,272 questions. Due to computational constraints associated with the 70b parameter model, we used a randomly sampled subset of 400 questions (MiniTest) to evaluate our model. We report the average accuracy. If no answer is output, this run will not be counted. This causes accuracy to not be multiples of 0.25 at 70b models. A comprehensive evaluation with the 8b model was conducted on the full development set to further validate our model. All experiments utilized Ollama\footnote{https://ollama.com/} as the model serving platform with default configurations and GGUF Q4\_K\_M (int4) quantization, simulating local LLM deployment in a hospital setting. In our experiment, we set $\alpha=0.6$ in Eq.~\eqref{eq:alpha}; this value is further examined in the sensitivity analysis below.

\subsection{Performance Comparison Analysis}

In this section, we evaluate the performance of ToT across LLaMA2-70b, LLaMA3-70b, Qwen2.5-72b on the MiniTest and LLaMA3.1-8b on MedQA-USMLE test set.

\begin{table}[tbp]
\caption{Accuracy of Quantized (int4) LLMs on MedQA-USMLE}
\centering
\footnotesize 
\begin{tabular}{lcccc}
\toprule
Models & CoT-AVG & CoT-SC & ToT & QM-ToT \\
\midrule
LLaMA2-70b & 34\% & 27.75\% & 35.25\% & \textbf{50.25\%} \\
LLaMA3-70b & 78.65\% & \textbf{80.00\%} & 76.72\% & \text{79.70\%} \\
Qwen2.5-72b & 72.37\% & 70.25\% & 73.00\% & \textbf{74.25\%} \\
LLaMA3.1-8b & 58.77\% & 59.19\% & 60.77\% & \textbf{69.49\%} \\
\bottomrule
\multicolumn{5}{l}{CoT-AVG represents the average accuracy achieved across all questions}\\
\multicolumn{5}{l}{when CoT reasoning is applied. “ToT”: QM-ToT framework without}\\
\multicolumn{5}{l}{the evaluator module.}\\
\end{tabular}
\label{tab:1}
\end{table}

\begin{figure}[!ht]
    \centering
    \includegraphics[width=1\linewidth]{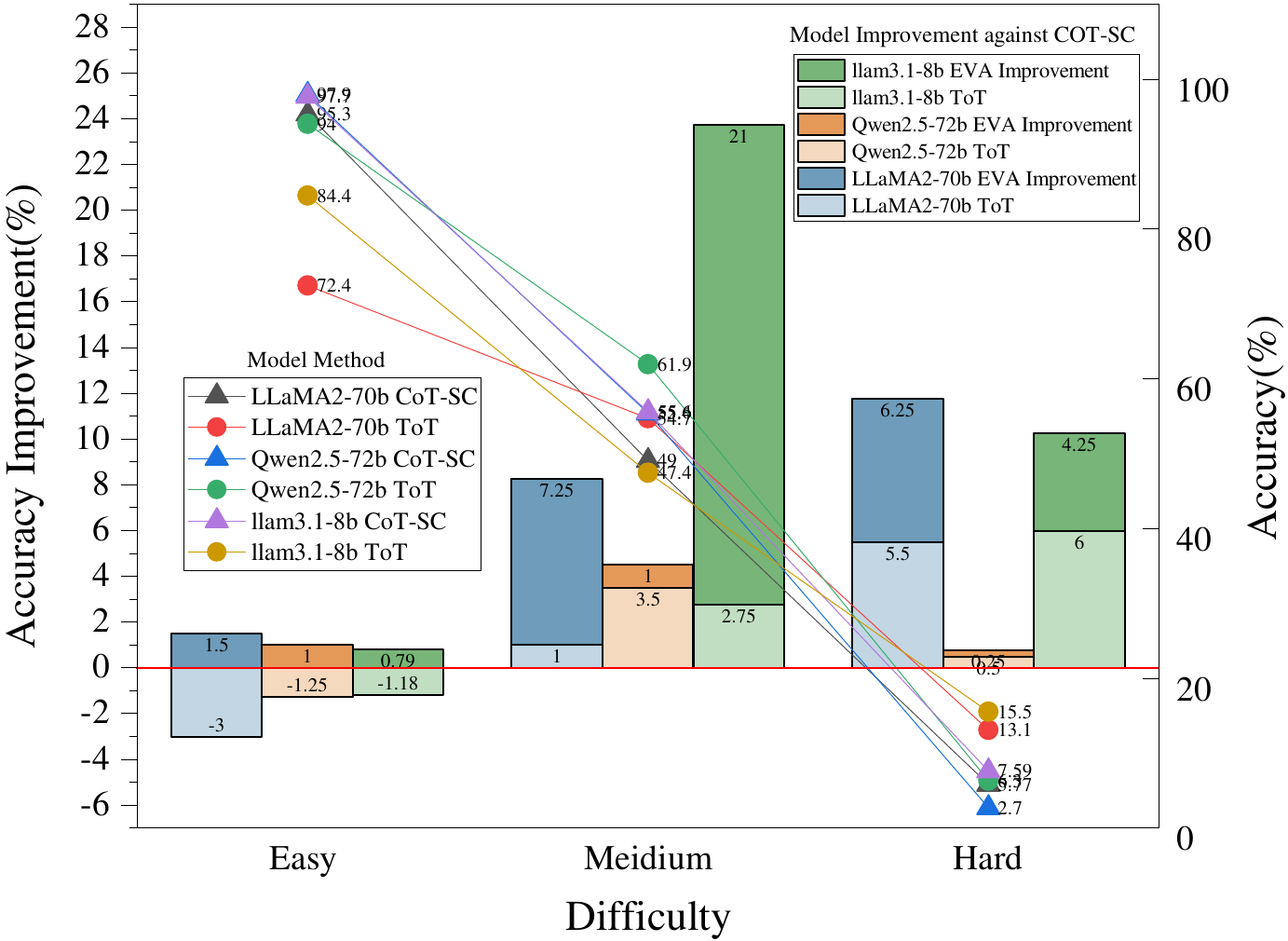}
    \caption{Performance of LLMs with CoT-SC and QM-ToT across difficulty levels. The line plot illustrates accuracy percentages in ToT and CoT-SC results for three models with values annotated. The ToT bars plot highlights accuracy improvements of ToT over CoT-SC, with a red dashed line indicating zero improvement. The EVA Improvement bars illustrate the accuracy improvement of ToT + EVA (i.e QM-ToT) compared to ToT.}
    \label{CoT-SCvsQM-ToT}
\end{figure}
\begin{figure}[!ht]
    \centering
    \includegraphics[width=1\linewidth]{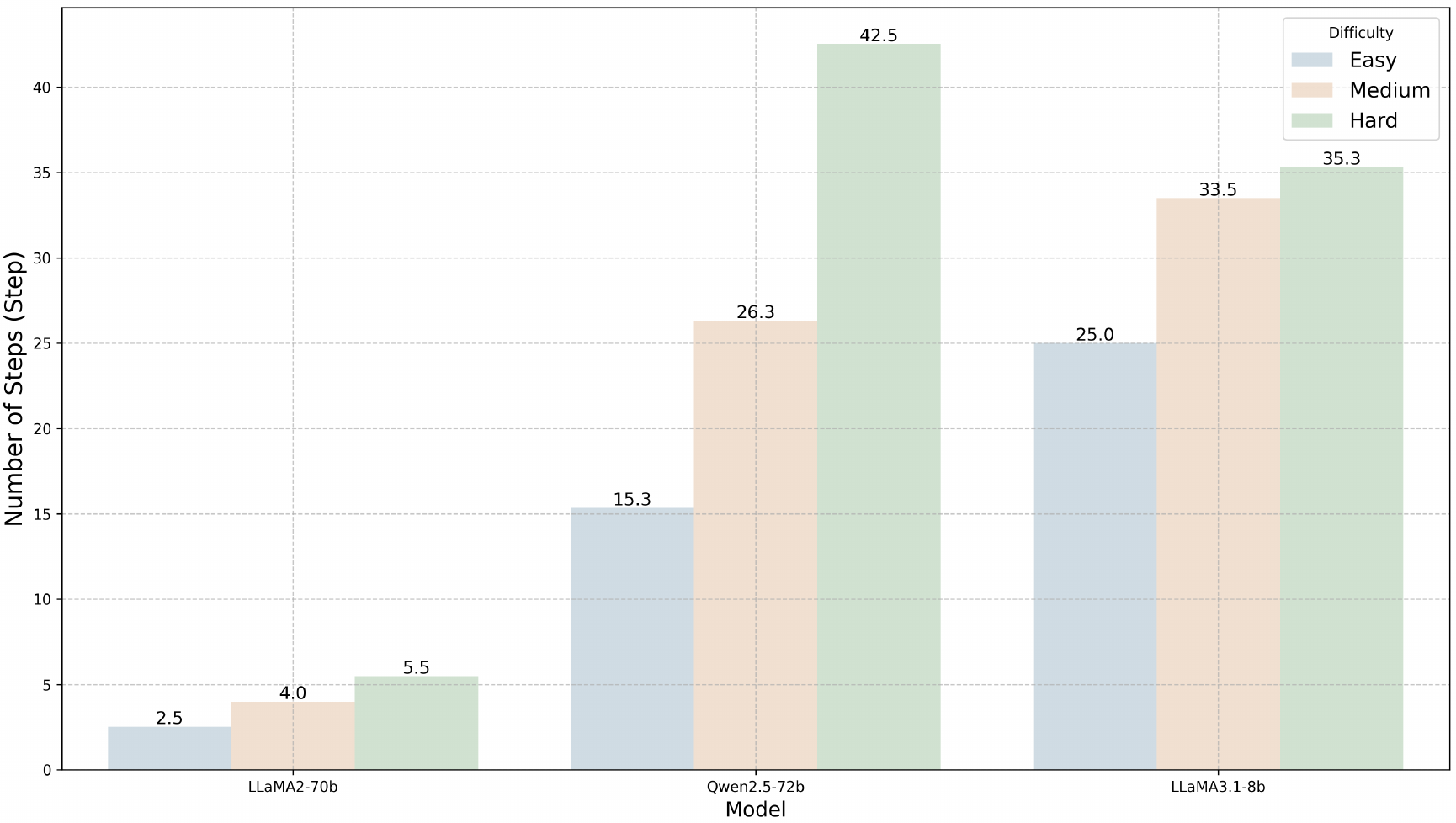}
    \caption{Average number of paths in different levels required by LLMs using QM-ToT.}
    \label{fig:5}
\end{figure}

TABLE~\ref{tab:1} presents the accuracy of quantized (int4) LLMs on the MedQA-USMLE dataset across different reasoning methods. It highlights the progressive improvement in accuracy from CoT-AVG to QM-ToT, showcasing the effectiveness of more sophisticated reasoning techniques. For instance, LLaMA2-70b improves from 34\% under CoT-AVG to 50.25\% with QM-ToT, demonstrating a significant gain. The results emphasize the consistent superiority of QM-ToT over simpler methods like CoT-SC and ToT, as seen across all models except LLaMA3-70b, including smaller ones like LLaMA3.1-8b, which improves from 58.77\% (CoT-AVG) to 69.49\% (QM-ToT). For LLaMA3-70B, QM-ToT obtains 79.70\%, which is close to CoT-SC at 80.00\%. Given the 400-question MiniTest size and the overlapping confidence intervals, this small difference should be treated as statistically indistinguishable rather than as evidence of degradation.
\begin{figure}
    \centering
    \includegraphics[width=1\linewidth]{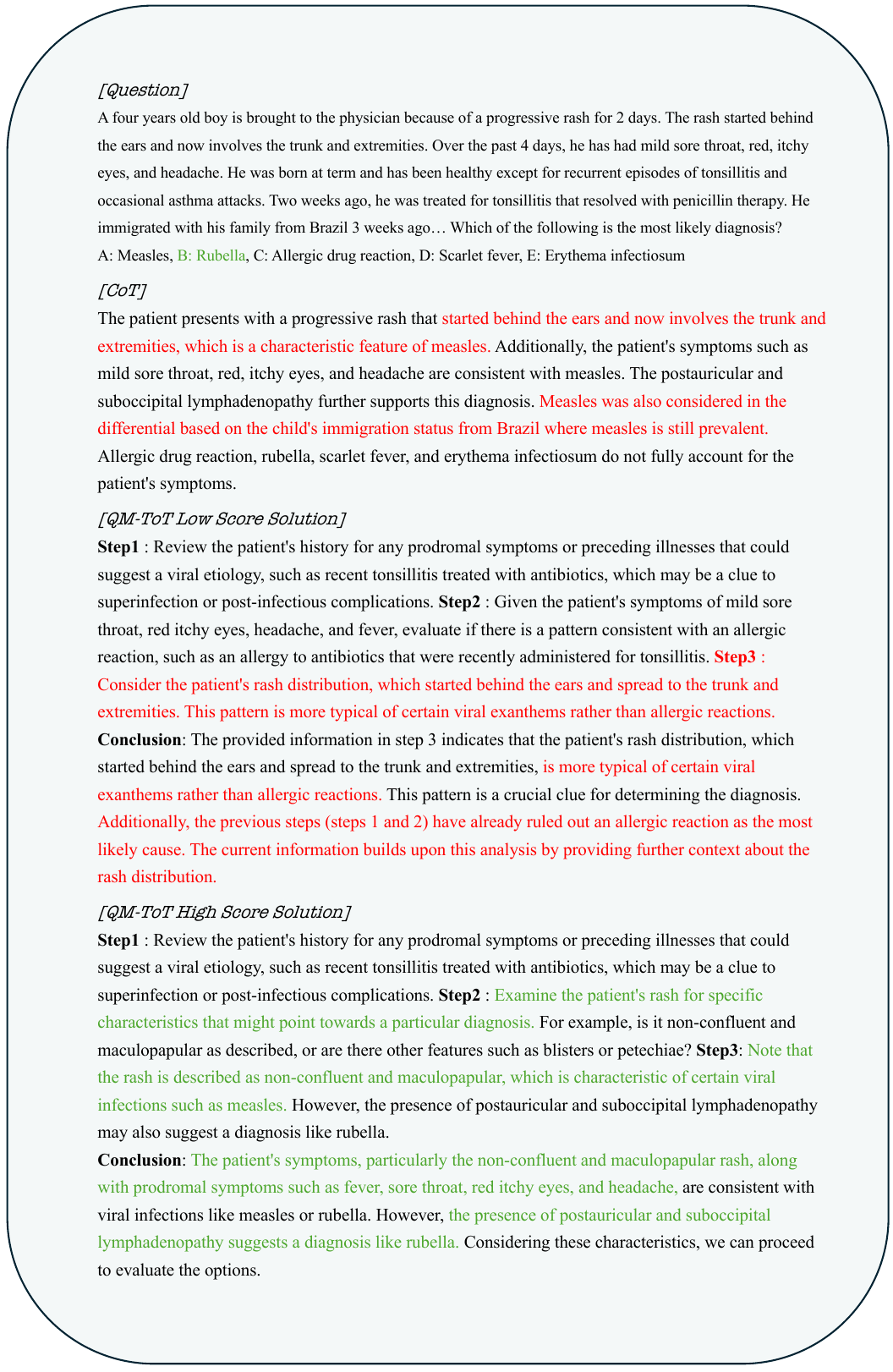}
    \caption{Example of different solution to the \#44 question. All the solutions are generated by LlaMA3.1-8b-INT4. The red text indicates the reasoning process that led to incorrect results, while the green text shows the reasoning process that led to correct results.}
    \label{fig:example}
\end{figure}

Fig.~\ref{fig:example} presents an example of different solutions to the same question. Compared to the CoT solution's misleading at the beginning, the ToT-generated solution is more detailed. Moreover, it is evident that even two different paths starting from the same point (step 1) may result in significantly different scores. The divergent directions taken in their subsequent steps lead to different outcomes. The low score solution focused on incorrect diagnostic features in step 3, which led to an incorrect solution. But This further demonstrates that the QM-ToT method can effectively break down problems and enable the model to approach problems from multiple perspectives.

\subsection{Effectiveness Analysis across Difficulty Levels}

We analyze the effectiveness of ToT across varying question difficulties, categorized as easy, medium, and hard based on the reason boundary described in Section~\ref{sec:rb}.  Our analysis examines answer distributions, accuracy improvements, and path length distributions.

\subsubsection{Analysis of CoT-SC and QM-ToT across Question Difficulties}
Fig.~\ref{CoT-SCvsQM-ToT} presents an analysis of CoT-SC, ToT and the improvement caused by evaluator modul (EVA Improvement) across three LLMs: LLaMA2-70b, Qwen2.5-72b, and LLaMA3-70b. Performance is evaluated on tasks of varying difficulty (easy, medium, and hard). The line graph illustrates the accuracy of each model-method pairing, revealing a consistent trend: ToT outperforms CoT-SC in medium and hard levels, but shows a decrease in the easy level. This is further corroborated by the ToT bar graph, which quantifies the improvement of ToT over CoT-SC. The EVA Improvement bars also show an improvement for all models with each difficulty level, which means the evaluator module can effectively evaluate ToT's solutions. LLaMA3.1-8b shows particularly strong performance gains with evaluator, especially on medium difficulty questions, achieving a 21\% improvement. For the whole QM-ToT workflow, while minimal or negative improvement is observed for easy questions, substantial gains are evident for medium and hard questions, suggesting QM-ToT's efficacy in complex reasoning.

\subsubsection{Autonomous Planning and Path Scaling}
Fig.~\ref{fig:5} shows that the average number of reasoning paths increases with question difficulty, indicating a strong correlation between task complexity and path density.
This adaptive behavior reflects the path-scaling property of QM-ToT. Rather than following a fixed reasoning pattern, the framework allocates more reasoning paths to complex clinical cases while requiring fewer paths for simpler queries.
These observations, together with the accuracy degradation shown in Fig.~\ref{fig:1}, 
indicate that while INT4 quantization reduces final performance, it does not disrupt the model’s ability to scale its reasoning paths with task difficulty. 
This suggests that structured path expansion is relatively robust to low-precision deployment. As a result, QM-ToT is a practical solution for reliable medical reasoning in resource-constrained settings.

\subsection{Ablation Study}
\subsubsection{Statistical reliability.}
To quantify the uncertainty caused by the MiniTest sample size, we report 95\% Clopper--Pearson confidence intervals for the QM-ToT results in Table~\ref{tab:qmtot_ci}. The 70B-class models are evaluated on 400 questions, leading to confidence intervals of approximately $\pm4$--$5$ percentage points. Therefore, the small difference between LLaMA3-70B with CoT-SC (80.00\%) and QM-ToT (79.70\%) should be interpreted as statistically indistinguishable rather than as a meaningful degradation.

\begin{table}
\caption{95\% confidence intervals for QM-ToT accuracy.}
\centering
\footnotesize
\begin{tabular}{lccc}
\toprule
Model & Accuracy & $n$ & 95\% CI \\
\midrule
LLaMA2-70B & 50.25\% & 400 & [45.24\%, 55.01\%] \\
LLaMA3-70B & 79.70\% & 400 & [75.47\%, 83.35\%] \\
Qwen2.5-72B & 74.25\% & 400 & [69.67\%, 78.23\%] \\
LLaMA3.1-8B & 69.49\% & 1272 & [66.88\%, 71.94\%] \\
\bottomrule
\end{tabular}
\label{tab:qmtot_ci}
\end{table}

\subsubsection{Sensitivity to $\alpha$.}
We further examine the weighting coefficient $\alpha$ in Eq.~\eqref{eq:alpha}, which balances reasoning coherence and medical correctness in the final evaluator score. As shown in Table~\ref{tab:alpha_sensitivity}, using both scores consistently outperforms relying on either score alone. The selected value $\alpha=0.6$ is close to the best setting for all models and remains within a stable high-performing range from 0.4 to 0.7.

\begin{table}
\caption{Accuracy under different $\alpha$ values.}
\centering
\footnotesize
\resizebox{\columnwidth}{!}{%
\begin{tabular}{lccccccc}
\toprule
Model & 0.0 & 0.2 & 0.4 & 0.5 & 0.6 & 0.8 & 1.0 \\
\midrule
LLaMA2-70B & 48.02\% & 49.09\% & 50.19\% & 49.75\% & \textbf{50.25\%} & 49.05\% & 47.60\% \\
LLaMA3-70B & 78.50\% & 79.80\% & 79.73\% & \textbf{80.10\%} & 79.70\% & 79.47\% & 78.28\% \\
Qwen2.5-72B & 72.01\% & 72.91\% & 73.52\% & 74.18\% & \textbf{74.25\%} & 73.40\% & 72.01\% \\
LLaMA3.1-8B & 68.56\% & 68.88\% & 68.20\% & 69.32\% & \textbf{69.49\%} & 68.15\% & 67.15\% \\
\bottomrule
\end{tabular}}
\label{tab:alpha_sensitivity}
\end{table}

\subsubsection{Two-stage answer selection.}
We also ablate the two-stage selection rule that compares the option with the highest average score and the option with the highest individual path score. Table~\ref{tab:selection_ablation} shows that the two-stage strategy consistently outperforms both Avg-Only and Score-Only selection. The disagreement cases are frequent (28\%--35\%), and the judge resolves them correctly in 62\%--75\% of cases, indicating that the second-stage decision is not redundant.

\begin{table}
\caption{Ablation of the two-stage answer selection strategy.}
\centering
\footnotesize
\resizebox{\columnwidth}{!}{%
\begin{tabular}{lcccc}
\toprule
Statistic & LLaMA2-70B & LLaMA3-70B & Qwen2.5-72B & LLaMA3.1-8B \\
\midrule
Agreement rate & 53\% & 72\% & 71\% & 68\% \\
Disagreement rate & 47\% & 28\% & 30\% & 32\% \\
Avg-Only accuracy & 43.55\% & 78.69\% & 73.54\% & 68.13\% \\
Score-Only accuracy & 45.67\% & 77.84\% & 72.60\% & 67.41\% \\
Two-Stage accuracy & \textbf{50.25\%} & \textbf{79.70\%} & \textbf{74.25\%} & \textbf{69.49\%} \\
Judge win rate & 62\% & 75\% & 72\% & 70\% \\
\bottomrule
\end{tabular}}
\label{tab:selection_ablation}
\end{table}

\subsubsection{FP16 control experiment.}
To separate the general effect of the QM-ToT reasoning structure from the quantization-specific effect, we additionally evaluate QM-ToT on the FP16 version of LLaMA3.1-8B. As shown in Table~\ref{tab:fp16_control}, QM-ToT also improves the FP16 model, but the gain is much smaller than that observed under INT4 quantization. This suggests that the tree-structured reasoning process is generally useful, while quantized models benefit more strongly from structured search and evaluator feedback.

\begin{table}
\caption{FP16 control experiment on LLaMA3.1-8B.}
\centering
\footnotesize
\begin{tabular}{lccc}
\toprule
Precision & CoT-SC & QM-ToT & Improvement \\
\midrule
FP16 & 66.43\% & \textbf{68.09\%} & +1.66\% \\
INT4 & 59.19\% & \textbf{69.49\%} & +10.30\% \\
\bottomrule
\end{tabular}
\label{tab:fp16_control}
\end{table}

\subsubsection{Self-evaluator baseline.}
Because QM-ToT uses DeepSeek-V3 as an external evaluator, we further replace it with the student model itself to isolate the contribution of tree-structured reasoning. Table~\ref{tab:self_eval} shows that Self-Eval QM-ToT improves over CoT-SC for LLaMA2-70B, Qwen2.5-72B, and LLaMA3.1-8B, indicating that the ToT structure itself contributes to performance. DeepSeek-V3 provides additional gains, especially for weaker models, confirming that both structured search and evaluator strength affect the final result.

\begin{table}
\caption{Self-evaluator baseline for QM-ToT.}
\centering
\footnotesize
\resizebox{\columnwidth}{!}{%
\begin{tabular}{lccccc}
\toprule
Model & CoT-SC & Self-Eval QM-ToT & DeepSeek QM-ToT & Self-Eval Gain & DeepSeek Extra Gain \\
\midrule
LLaMA2-70B & 27.75\% & 38.50\% & \textbf{50.25\%} & +10.75\% & +11.75\% \\
LLaMA3-70B & 80.00\% & 78.25\% & \textbf{79.70\%} & -1.75\% & +1.70\% \\
Qwen2.5-72B & 70.25\% & 71.50\% & \textbf{74.25\%} & +1.25\% & +2.80\% \\
LLaMA3.1-8B & 59.19\% & 63.11\% & \textbf{69.49\%} &  +3.92\% & +6.38\% \\
\bottomrule
\end{tabular}}
\label{tab:self_eval}
\end{table}

\subsubsection{CoT-SC with DeepSeek selector.}
We also test whether the improvement can be explained solely by using DeepSeek-V3 as a selector. In this baseline, CoT-SC generates multiple reasoning chains and DeepSeek-V3 selects the best answer without the QM-ToT tree-search process. Table~\ref{tab:deepseek_selector} shows that CoT-SC with DeepSeek improves over plain CoT-SC in most cases, but remains below full QM-ToT. This gap indicates that the tree-structured exploration provides additional benefit beyond external answer selection alone.

\begin{table}
\caption{CoT-SC with DeepSeek selector baseline.}
\centering
\footnotesize
\resizebox{\columnwidth}{!}{%
\begin{tabular}{lcccc}
\toprule
Model & CoT-SC & CoT-SC + DeepSeek & QM-ToT & ToT Extra Gain \\
\midrule
LLaMA2-70B & 27.75\% & 42.00\% & \textbf{50.25\%} & +8.25\% \\
LLaMA3-70B & 80.00\% & 78.50\% & \textbf{79.70\%} & +1.20\% \\
Qwen2.5-72B & 70.25\% & 72.50\% & \textbf{74.25\%} & +1.75\% \\
LLaMA3.1-8B & 59.19\% & 64.32\% & \textbf{69.49\%} & +5.17\% \\
\bottomrule
\end{tabular}}
\label{tab:deepseek_selector}
\end{table}

\subsubsection{Multi-seed stability.}
To assess robustness to sampling variation, we repeat QM-ToT with three random seeds. As shown in Table~\ref{tab:seed_stability}, the standard deviation remains around 1 percentage point across all models, suggesting that the observed QM-ToT performance is stable across seeds.

\begin{table}
\caption{QM-ToT accuracy across three random seeds.}
\centering
\footnotesize
\begin{tabular}{lcccc}
\toprule
Model & Seed 1 & Seed 2 & Seed 3 & Mean $\pm$ Std \\
\midrule
LLaMA2-70B & 51.25\% & 48.50\% & 51.00\% & \textbf{49.90 $\pm$ 1.30\%} \\
LLaMA3-70B & 80.10\% & 80.50\% & 78.50\% & \textbf{79.60 $\pm$ 1.00\%} \\
Qwen2.5-72B & 74.25\% & 73.00\% & 75.50\% & \textbf{74.30 $\pm$ 1.30\%} \\
LLaMA3.1-8B & 69.82\% & 68.47\% & 70.18\% & \textbf{69.40 $\pm$ 0.90\%} \\
\bottomrule
\end{tabular}
\label{tab:seed_stability}
\end{table}

\begin{figure}
    \centering
    \includegraphics[width=0.7\linewidth]{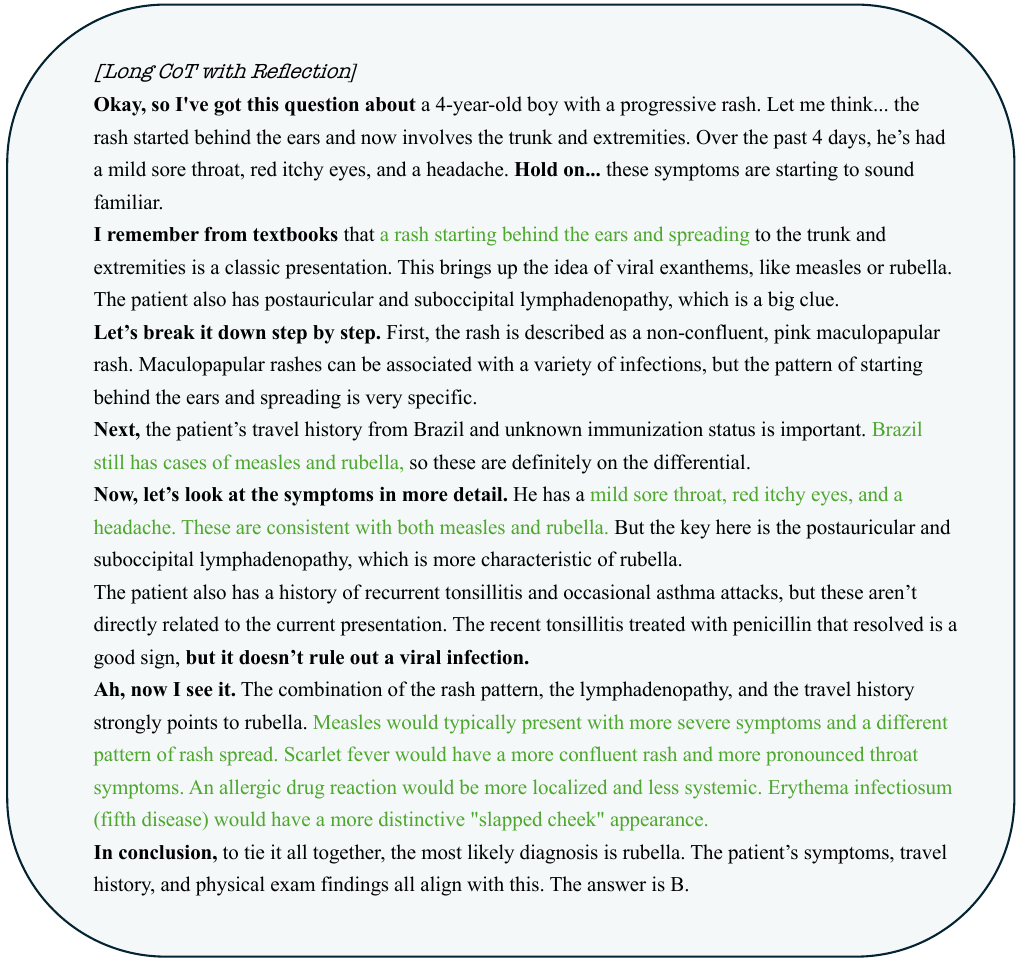}
    \caption{Example of right long CoT to the \#44 question from Reflection-ToT. The bold text demonstrate the reflection thinking process. The green text shows the reasoning process that led to correct results.}
    \label{fig:Reflection-ToT-example}
\end{figure}
\subsection{Performance of Reflection-ToT Distillation}
In this section, we use Llamafactory~\cite{llamafactory} as the training framework to measure LoRA~\cite{lora} adapter performance on LLaMA3.1-8B for MedQA-USMLE-test. We use MedQA-USMLE-dev and MedQA-USMLE-train as the training set for distillation data. For the DPO training, we use LLaMA3-70B as the teacher model and set the learning rate to 5e-5, train batch size to 16, and cut-off length to 3k, training for one epoch with 1,000 randomly sampled questions from the training set. For SFT training, we use QwQ as the teacher model and keep the same training hyperparameters.
After Reflection-ToT, short CoT generated by ToT is refined into extended o1-style CoT with reflection and summarization, as shown in Fig.~\ref{fig:Reflection-ToT-example}. Table~\ref{tab:2} compares Reflection-ToT with two QwQ-based CoT settings. Full QwQ-CoT distillation uses 10,178 questions and reaches 65.01\% accuracy, while a QwQ-based matched-pair setting with 1,514 DPO pairs reaches 64.73\%. Reflection-ToT uses only 1,000 questions to construct 1,514 preference pairs and achieves 66.44\% accuracy. These results suggest that Reflection-ToT can produce competitive distillation data with fewer source questions, but the comparison should be interpreted cautiously because the teacher model and data construction process differ across settings.

\begin{table}[t]
\caption{Comparison of Reflection-ToT and CoT Approaches}
\centering
\footnotesize 
\resizebox{\columnwidth}{!}{%
\begin{tabular}{lcccc}
\toprule
 & Baseline & CoT & CoT (Matched Pairs) & Reflection-ToT\\
\midrule
Teacher Model & \text{-} & QwQ & QwQ & LLaMA3-70B \\
Student Model & LLaMA3.1-8b & LLaMA3.1-8b & LLaMA3.1-8b & LLaMA3.1-8b \\
Question Used $\downarrow$ & \text{-} & 10178 & 10178 & 1000 \\
Generated Data $\downarrow$ & \text{-} & 6659 (samples) & 1514(pairs) & 1514 (pairs) \\
Accuracy$\uparrow$ & 61.01\% & 65.01\% & 64.73\% & \textbf{66.44\%} \\
\bottomrule
\end{tabular}
}
\label{tab:2}
\end{table}

\section{Limitations}

First, QM-ToT improves medical question-answering accuracy by expanding and evaluating multiple reasoning paths, but this design increases inference cost compared with CoT-SC. Based on the average number of paths and prompt length, QM-ToT requires substantially more tokens, with the estimated cost ranging from about $2.8\times$ to $20.5\times$ that of CoT-SC depending on model and question difficulty. Therefore, although QM-ToT is useful for improving reliability under quantized deployment, future work should optimize path pruning, early stopping, and evaluator invocation to reduce token and latency overhead.

Second, QM-ToT depends on an external evaluator. Although the self-evaluator and DeepSeek-selector ablations indicate that both tree-structured search and evaluator quality contribute to performance, the framework is not a purely self-contained reasoning method. Future work should further isolate evaluator knowledge from search-structure benefits across more model families.

Third, our difficulty stratification is based on CoT-SC accuracy. This makes the easy, medium, and hard analysis useful for diagnosing where CoT-SC fails, but it should not be interpreted as an independent validation of task difficulty. Future studies should include external difficulty criteria such as question category, USMLE step, expert annotation, or model-independent item statistics.

Finally, the 70B-class model results are evaluated on a 400-question MiniTest subset. We report confidence intervals and multi-seed results to quantify this uncertainty, but small differences of a few percentage points should still be interpreted cautiously.
\section{Conclusion}
In this paper, we introduced QM-ToT, a structured tree-of-thought reasoning framework that enables feedback-guided clinical reasoning in quantized LLMs. Our approach demonstrates that structured reasoning with evaluator feedback can help bridge the gap between model efficiency and reasoning reliability in healthcare. At the same time, this reliability gain comes with a higher inference-token cost, which should be considered when deploying QM-ToT in budget-constrained clinical environments. The framework's ability to dynamically allocate reasoning resources based on task complexity, as evidenced by the path scaling analysis, highlights its robustness and practicality for deployment in resource-constrained medical settings. Importantly, our method's ability to generate consistent, high-quality solutions offers a promising avenue for synthesizing domain-specific training data, potentially enabling more effective and intrinsically aligned post-training strategies. This work provides a viable solution for deploying LLMs in resource-constrained biomedical environments by reducing reliance on large-scale proprietary models and a ToT-driven low-cost data distillation strategy for medical tasks.

Future research directions include integrating advanced search strategies like Monte Carlo tree search~\cite{MCTS} and incorporating reinforcement learning~\cite{RLHF} techniques to further optimize the ToT framework. Other reasoning enhancement strategies such as neural architecture search~\cite{chen1}, black-box optimization~\cite{chen2} and evolutionary computation~\cite{chen3} could also be used for optimization. Leveraging the rich, chain-of-thought solution data generated by our method for fine-tuning LLMs presents another promising opportunity to enhance problem-solving capabilities. Moreover, the extended reasoning chains produced by ToT could facilitate the development of ``deep thinking" models specifically tailored for the biomedical domain, paving the way for more sophisticated and reliable AI-driven solutions in healthcare.

\section*{Acknowledgment}
The authors disclose the use of Claude and GPT~\cite{claude,chatgpt} models solely for language editing and proofreading of this manuscript.
No AI system was used for idea generation, experiment design, data analysis, or result interpretation.
The authors take full responsibility for the content of this paper.

%
%
%
\bibliographystyle{splncs04}
%
\bibliography{references}  
\end{document}